\batchmode
\makeatletter
\def\input@path{{/home/fyzhu/DATA/Dropbox/self_Folder/myWorksOnDropboxs/201702_MICCAI_RobustRL_4_mHealth_onlineLearning//}}
\makeatother
\documentclass{llncs}
\usepackage[latin9]{inputenc}
\usepackage{color}
\definecolor{page_backgroundcolor}{rgb}{1, 1, 1}
\pagecolor{page_backgroundcolor}
\usepackage{array}
\usepackage{float}
\usepackage{bm}
\usepackage{multirow}
\usepackage{amsmath}
\usepackage{amssymb}
\usepackage{graphicx}
\usepackage[unicode=true,pdfusetitle,
 bookmarks=true,bookmarksnumbered=false,bookmarksopen=false,
 breaklinks=true,pdfborder={0 0 0},backref=false,colorlinks=true]
 {hyperref}

\makeatletter

\providecommand{\tabularnewline}{\\}
\floatstyle{ruled}
\newfloat{algorithm}{tbp}{loa}
\providecommand{\algorithmname}{Algorithm}
\floatname{algorithm}{\protect\algorithmname}

\usepackage{llncsdoc}
\usepackage{makeidx}\usepackage{epsfig}% allows for indexgeneration
\usepackage{url}
\usepackage{epsfig}
\usepackage{algorithm}
\usepackage{algorithmic}
\usepackage{subfigure}
\usepackage{caption}
\usepackage{epsfig}
\usepackage{epstopdf}
\usepackage{adjustbox}

\makeatother

\begin{document}
\global\long\def\mtbfA{\mathbf{A}}
 \global\long\def\mtbfa{\mathbf{a}}
 \global\long\def\mebfA{\bar{\mtbfA}}
 \global\long\def\mebfa{\bar{\mtbfa}}

\global\long\def\mhbfA{\widehat{\mathbf{A}}}
 \global\long\def\mhbfa{\widehat{\mathbf{a}}}
 \global\long\def\mtcalA{\mathcal{A}}
 \global\long\def\mtbbA{\mathbb{A}}

\global\long\def\mtbfB{\mathbf{B}}
 \global\long\def\mtbfb{\mathbf{b}}
 \global\long\def\mebfB{\bar{\mtbfB}}
 \global\long\def\mebfb{\bar{\mtbfb}}

\global\long\def\mhbfB{\widehat{\mathbf{B}}}
 \global\long\def\mhbfb{\widehat{\mathbf{b}}}
 \global\long\def\mtcalB{\mathcal{B}}
 \global\long\def\mtbbB{\mathbb{B}}

\global\long\def\mtbfC{\mathbf{C}}
 \global\long\def\mtbfc{\mathbf{c}}
 \global\long\def\mebfC{\bar{\mtbfC}}
 \global\long\def\mebfc{\bar{\mtbfc}}

\global\long\def\mhbfC{\widehat{\mathbf{C}}}
 \global\long\def\mhbfc{\widehat{\mathbf{c}}}
 \global\long\def\mtcalC{\mathcal{C}}
 \global\long\def\mtbbC{\mathbb{C}}

\global\long\def\mtbfD{\mathbf{D}}
 \global\long\def\mtbfd{\mathbf{d}}
 \global\long\def\mebfD{\bar{\mtbfD}}
 \global\long\def\mebfd{\bar{\mtbfd}}

\global\long\def\mhbfD{\widehat{\mathbf{D}}}
 \global\long\def\mhbfd{\widehat{\mathbf{d}}}
 \global\long\def\mtcalD{\mathcal{D}}
 \global\long\def\mtbbD{\mathbb{D}}

\global\long\def\mtbfE{\mathbf{E}}
 \global\long\def\mtbfe{\mathbf{e}}
 \global\long\def\mebfE{\bar{\mtbfE}}
 \global\long\def\mebfe{\bar{\mtbfe}}

\global\long\def\mhbfE{\widehat{\mathbf{E}}}
 \global\long\def\mhbfe{\widehat{\mathbf{e}}}
 \global\long\def\mtcalE{\mathcal{E}}
 \global\long\def\mtbbE{\mathbb{E}}

\global\long\def\mtbfF{\mathbf{F}}
 \global\long\def\mtbff{\mathbf{f}}
 \global\long\def\mebfF{\bar{\mathbf{F}}}
 \global\long\def\mebff{\bar{\mathbf{f}}}

\global\long\def\mhbfF{\widehat{\mathbf{F}}}
 \global\long\def\mhbff{\widehat{\mathbf{f}}}
 \global\long\def\mtcalF{\mathcal{F}}
 \global\long\def\mtbbF{\mathbb{F}}

\global\long\def\mtbfG{\mathbf{G}}
 \global\long\def\mtbfg{\mathbf{g}}
 \global\long\def\mebfG{\bar{\mathbf{G}}}
 \global\long\def\mebfg{\bar{\mathbf{g}}}

\global\long\def\mhbfG{\widehat{\mathbf{G}}}
 \global\long\def\mhbfg{\widehat{\mathbf{g}}}
 \global\long\def\mtcalG{\mathcal{G}}
 \global\long\def\mtbbG{\mathbb{G}}

\global\long\def\mtbfH{\mathbf{H}}
 \global\long\def\mtbfh{\mathbf{h}}
 \global\long\def\mebfH{\bar{\mathbf{H}}}
 \global\long\def\mebfh{\bar{\mathbf{h}}}

\global\long\def\mhbfH{\widehat{\mathbf{H}}}
 \global\long\def\mhbfh{\widehat{\mathbf{h}}}
 \global\long\def\mtcalH{\mathcal{H}}
 \global\long\def\mtbbH{\mathbb{H}}

\global\long\def\mtbfI{\mathbf{I}}
 \global\long\def\mtbfi{\mathbf{i}}
 \global\long\def\mebfI{\bar{\mathbf{I}}}
 \global\long\def\mebfi{\bar{\mathbf{i}}}

\global\long\def\mhbfI{\widehat{\mathbf{I}}}
 \global\long\def\mhbfi{\widehat{\mathbf{i}}}
 \global\long\def\mtcalI{\mathcal{I}}
 \global\long\def\mtbbI{\mathbb{I}}

\global\long\def\mtbfJ{\mathbf{J}}
 \global\long\def\mtbfj{\mathbf{j}}
 \global\long\def\mebfJ{\bar{\mathbf{J}}}
 \global\long\def\mebfj{\bar{\mathbf{j}}}

\global\long\def\mhbfJ{\widehat{\mathbf{J}}}
 \global\long\def\mhbfj{\widehat{\mathbf{j}}}
 \global\long\def\mtcalJ{\mathcal{J}}
 \global\long\def\mtbbJ{\mathbb{J}}

\global\long\def\mtbfK{\mathbf{K}}
 \global\long\def\mtbfk{\mathbf{k}}
 \global\long\def\mebfK{\bar{\mathbf{K}}}
 \global\long\def\mebfk{\bar{\mathbf{k}}}

\global\long\def\mhbfK{\widehat{\mathbf{K}}}
 \global\long\def\mhbfk{\widehat{\mathbf{k}}}
 \global\long\def\mtcalK{\mathcal{K}}
 \global\long\def\mtbbK{\mathbb{K}}

\global\long\def\mtbfL{\mathbf{L}}
 \global\long\def\mtbfl{\mathbf{l}}
 \global\long\def\mebfL{\bar{\mathbf{L}}}
 \global\long\def\mebfl{\bar{\mathbf{l}}}

\global\long\def\mhbfL{\widehat{\mathbf{K}}}
 \global\long\def\mhbfl{\widehat{\mathbf{k}}}
 \global\long\def\mtcalL{\mathcal{L}}
 \global\long\def\mtbbL{\mathbb{L}}

\global\long\def\mtbfM{\mathbf{M}}
 \global\long\def\mtbfm{\mathbf{m}}
 \global\long\def\mebfM{\bar{\mathbf{M}}}
 \global\long\def\mebfm{\bar{\mathbf{m}}}

\global\long\def\mhbfM{\widehat{\mathbf{M}}}
 \global\long\def\mhbfm{\widehat{\mathbf{m}}}
 \global\long\def\mtcalM{\mathcal{M}}
 \global\long\def\mtbbM{\mathbb{M}}

\global\long\def\mtbfN{\mathbf{N}}
 \global\long\def\mtbfn{\mathbf{n}}
 \global\long\def\mebfN{\bar{\mathbf{N}}}
 \global\long\def\mebfn{\bar{\mathbf{n}}}

\global\long\def\mhbfN{\widehat{\mathbf{N}}}
 \global\long\def\mhbfn{\widehat{\mathbf{n}}}
 \global\long\def\mtcalN{\mathcal{N}}
 \global\long\def\mtbbN{\mathbb{N}}

\global\long\def\mtbfO{\mathbf{O}}
 \global\long\def\mtbfo{\mathbf{o}}
 \global\long\def\mebfO{\bar{\mathbf{O}}}
 \global\long\def\mebfo{\bar{\mathbf{o}}}

\global\long\def\mhbfO{\widehat{\mathbf{O}}}
 \global\long\def\mhbfo{\widehat{\mathbf{o}}}
 \global\long\def\mtcalO{\mathcal{O}}
 \global\long\def\mtbbO{\mathbb{O}}

\global\long\def\mtbfP{\mathbf{P}}
 \global\long\def\mtbfp{\mathbf{p}}
 \global\long\def\mebfP{\bar{\mathbf{P}}}
 \global\long\def\mebfp{\bar{\mathbf{p}}}

\global\long\def\mhbfP{\widehat{\mathbf{P}}}
 \global\long\def\mhbfp{\widehat{\mathbf{p}}}
 \global\long\def\mtcalP{\mathcal{P}}
 \global\long\def\mtbbP{\mathbb{P}}

\global\long\def\mtbfQ{\mathbf{Q}}
 \global\long\def\mtbfq{\mathbf{q}}
 \global\long\def\mebfQ{\bar{\mathbf{Q}}}
 \global\long\def\mebfq{\bar{\mathbf{q}}}

\global\long\def\mhbfQ{\widehat{\mathbf{Q}}}
 \global\long\def\mhbfq{\widehat{\mathbf{q}}}
\global\long\def\mtcalQ{\mathcal{Q}}
 \global\long\def\mtbbQ{\mathbb{Q}}

\global\long\def\mtbfR{\mathbf{R}}
 \global\long\def\mtbfr{\mathbf{r}}
 \global\long\def\mebfR{\bar{\mathbf{R}}}
 \global\long\def\mebfr{\bar{\mathbf{r}}}

\global\long\def\mhbfR{\widehat{\mathbf{R}}}
 \global\long\def\mhbfr{\widehat{\mathbf{r}}}
\global\long\def\mtcalR{\mathcal{R}}
 \global\long\def\mtbbR{\mathbb{R}}

\global\long\def\mtbfS{\mathbf{S}}
 \global\long\def\mtbfs{\mathbf{s}}
 \global\long\def\mebfS{\bar{\mathbf{S}}}
 \global\long\def\mebfs{\bar{\mathbf{s}}}

\global\long\def\mhbfS{\widehat{\mathbf{S}}}
 \global\long\def\mhbfs{\widehat{\mathbf{s}}}
\global\long\def\mtcalS{\mathcal{S}}
 \global\long\def\mtbbS{\mathbb{S}}

\global\long\def\mtbfT{\mathbf{T}}
 \global\long\def\mtbft{\mathbf{t}}
 \global\long\def\mebfT{\bar{\mathbf{T}}}
 \global\long\def\mebft{\bar{\mathbf{t}}}

\global\long\def\mhbfT{\widehat{\mathbf{T}}}
 \global\long\def\mhbft{\widehat{\mathbf{t}}}
 \global\long\def\mtcalT{\mathcal{T}}
 \global\long\def\mtbbT{\mathbb{T}}

\global\long\def\mtbfU{\mathbf{U}}
 \global\long\def\mtbfu{\mathbf{u}}
 \global\long\def\mebfU{\bar{\mathbf{U}}}
 \global\long\def\mebfu{\bar{\mathbf{u}}}

\global\long\def\mhbfU{\widehat{\mathbf{U}}}
 \global\long\def\mhbfu{\widehat{\mathbf{u}}}
 \global\long\def\mtcalU{\mathcal{U}}
 \global\long\def\mtbbU{\mathbb{U}}

\global\long\def\mtbfV{\mathbf{V}}
 \global\long\def\mtbfv{\mathbf{v}}
 \global\long\def\mebfV{\bar{\mathbf{V}}}
 \global\long\def\mebfv{\bar{\mathbf{v}}}

\global\long\def\mhbfV{\widehat{\mathbf{V}}}
 \global\long\def\mhbfv{\widehat{\mathbf{v}}}
\global\long\def\mtcalV{\mathcal{V}}
 \global\long\def\mtbbV{\mathbb{V}}

\global\long\def\mtbfW{\mathbf{W}}
 \global\long\def\mtbfw{\mathbf{w}}
 \global\long\def\mebfW{\bar{\mathbf{W}}}
 \global\long\def\mebfw{\bar{\mathbf{w}}}

\global\long\def\mhbfW{\widehat{\mathbf{W}}}
 \global\long\def\mhbfw{\widehat{\mathbf{w}}}
 \global\long\def\mtcalW{\mathcal{W}}
 \global\long\def\mtbbW{\mathbb{W}}

\global\long\def\mtbfX{\mathbf{X}}
 \global\long\def\mtbfx{\mathbf{x}}
 \global\long\def\mebfX{\bar{\mtbfX}}
 \global\long\def\mebfx{\bar{\mtbfx}}

\global\long\def\mhbfX{\widehat{\mathbf{X}}}
 \global\long\def\mhbfx{\widehat{\mathbf{x}}}
 \global\long\def\mtcalX{\mathcal{X}}
 \global\long\def\mtbbX{\mathbb{X}}

\global\long\def\mtbfY{\mathbf{Y}}
 \global\long\def\mtbfy{\mathbf{y}}
\global\long\def\mebfY{\bar{\mathbf{Y}}}
 \global\long\def\mebfy{\bar{\mathbf{y}}}

\global\long\def\mhbfY{\widehat{\mathbf{Y}}}
 \global\long\def\mhbfy{\widehat{\mathbf{y}}}
 \global\long\def\mtcalY{\mathcal{Y}}
 \global\long\def\mtbbY{\mathbb{Y}}

\global\long\def\mtbfZ{\mathbf{Z}}
 \global\long\def\mtbfz{\mathbf{z}}
 \global\long\def\mebfZ{\bar{\mathbf{Z}}}
 \global\long\def\mebfz{\bar{\mathbf{z}}}

\global\long\def\mhbfZ{\widehat{\mathbf{Z}}}
 \global\long\def\mhbfz{\widehat{\mathbf{z}}}
\global\long\def\mtcalZ{\mathcal{Z}}
 \global\long\def\mtbbZ{\mathbb{Z}}

\global\long\def\mtth{\text{th}}

\global\long\def\mtbfzero{\mathbf{0}}
 \global\long\def\mtbfone{\mathbf{1}}

\global\long\def\mttrace{\text{Tr}}

\global\long\def\mttotalVariation{\text{TV}}

\global\long\def\mtexpect{\mathbb{E}}

\global\long\def\mtdet{\text{det}}

\global\long\def\mtvec{\mathbf{\text{vec}}}

\global\long\def\mtvar{\mathbf{\text{var}}}

\global\long\def\mtcov{\mathbf{\text{cov}}}

\global\long\def\mtsubTo{\mathbf{\text{s.t.}}}

\global\long\def\mtfor{\text{for}}

\global\long\def\mtrank{\text{rank}}

\global\long\def\mtrankn{\text{rankn}}

\global\long\def\mtdiag{\mathbf{\text{diag}}}

\global\long\def\mtsign{\mathbf{\text{sign}}}

\global\long\def\mtloss{\mathbf{\text{loss}}}

\global\long\def\mtwhen{\text{when}}

\global\long\def\mtwhere{\text{where}}

\global\long\def\mtif{\text{if}}

\title{Robust Contextual Bandit via the Capped-$\ell_{2}$ norm}

\author{Feiyun Zhu, Xinliang Zhu, Sheng Wang, Jiawen Yao, Junzhou Huang}

\institute{Department of CSE, University of Texas at Arlington, TX, 76013, USA}
\maketitle
\begin{abstract}
This paper considers the actor-critic contextual bandit for the mobile
health (mHealth) intervention. The state-of-the-art decision-making
methods in mHealth generally assume that the noise in the dynamic
system follows the Gaussian distribution. Those methods use the least-square-based
algorithm to estimate the expected reward, which is prone to the existence
of outliers. To deal with the issue of outliers, we propose a novel
robust actor-critic contextual bandit method for the mHealth intervention.
In the critic updating, the capped-$\ell_{2}$ norm is used to measure
the approximation error, which prevents outliers from dominating our
objective. A set of weights could be achieved from the critic updating.
Considering them gives a weighted objective for the actor updating.
It provides the badly noised sample in the critic updating with zero
weights for the actor updating. As a result, the robustness of both
actor-critic updating is enhanced. There is a key parameter in the
capped-$\ell_{2}$ norm. We provide a reliable method to properly
set it by making use of one of the most fundamental definitions of
outliers in statistics. Extensive experiment results demonstrate that
our method can achieve almost identical results compared with the
state-of-the-art methods on the dataset without outliers and dramatically
outperform them on the datasets noised by outliers.
\end{abstract}

\section{Introduction}

Nowadays, billions of people frequently use various kinds of smart
devices, such as smartphones and wearable activity sensors\ \cite{PengLiao_2015_Proposal_offPolicyRL,SusanMurphy_2016_CORR_BatchOffPolicyAvgRwd,fyZhu_2017_arXiv_CohesionDrivenActorCriticRL,fyzhu_2017_RLDM_WarmStart}.
It is increasingly popular among the scientist community to make use
of the state-of-the-art artificial intelligence technology to leverage
supercomputers and big data to facilicate the prediction of healthcare
tasks\ \cite{yaoyao_2017_MICCAI,xinliang_2017_CVPR_WSISA,zhengxu_2017_ACMBCB}.
In this paper, we use the mobile health (mHealth) technologies to
collect and analyze real-time data from users. Based on that, the
goal of mHealth is to decide when, where, and how to deliver the in-time
intervention to best serve uses, helping them to lead healthier lives.
For example, the mHealth guides people how to reduce alcohol abuses,
increase physical activities and regain the control of eating disorders,
obesity/weight management\ \cite{PengLiao_2015_Proposal_offPolicyRL,SusanMurphy_2016_CORR_BatchOffPolicyAvgRwd,huitian_2016_PhdThesis_actCriticAlgorithm}.

The tailoring of mHealth interventions is generally modeled as a sequential
decision-making (SDM) problem. The contextual bandit provides a paradigm
for the SDM\ \cite{Ambuj_2017_Springer_FromAds_Interventions,Zhou_2016_IJCAI_LatentContextualBandit,fyzhu_2017_RLDM_WarmStart,fyZhu_2017_arXiv_CohesionDrivenActorCriticRL}.
In mHealth, the first contextual bandit\ \cite{huitian_2014_NIPS_ActCriticBandit4JITAI}
was proposed in 2014. It is in an actor-critic setting and has an
explicit parameterized stochastic policy. Such setting has two advantages:
(1) the actor-critic algorithm has good properties of quick convergence
with low variance\ \cite{Grondman_2012_IEEEts_surveyOfActorCriticRL};
(2) we could understand the key features that contribute most to the
policy by analyzing the estimated parameters. This is important for
the behavior scientists to design the state (feature). Then, Dr. Lei\ \cite{huitian_2016_PhdThesis_actCriticAlgorithm}
improved the method by emphasizing the explorations and introducing
the stochasticity constraint on the policy coefficients. 

Those two methods serve a good start for the mHealth. However, they
assume that there is no outlier in the data. They use the least-square-based
algorithm to learn the expected reward, which, however, is prone to
the presence of outliers\ \cite{fyzhu_2014_AAAI_ARSS,fyzhu_2014_JSTSP_RRLbS,fyzhu_2015_PhDthesis,yingWang_2015_TIP_RobustUnmixing,guangliangCheng_2016_JStars_robustHyperClassification}.
In practice, there are various kinds of complex noise in the mHealth
system. For example, the wearable devices are unable to accurately
record the states and rewards from users under various conditions.
The mHealth requires self-report to deliver effective interventions
to device users. However, some users are unwilling to report the self-report.
They sometimes randomly fill out the report to save time. We treat
the various of complex noises in the system as outliers. We want to
get rid of the extreme observations. 

In this paper, a novel robust actor-critic contextual bandit is proposed
to deal with the outlier issue in the mHealth system. The capped-$\ell_{2}$
norm is used in the estimation of the expected reward in the critic
updating. As a result, we obtain a set of weights. With them, we propose
a weighted objective for the actor updating, which gives the samples
that are ineffective for the critic updating zero weights. As a result,
the robustness of both actor-critic updating is greatly enhanced.
There is a key parameter in the capped-$\ell_{2}$ norm. We propose
a solid method to set it properly, which is based on a solid method
to detect outliers in  statistics. With it, we can achieve the conflicting
goal of enhancing the robustness of our algorithm and obtaining almost
same results compared with the state-of-the-art method on the datasets
without outliers. Extensive experiment results show that in a variety
of parameter settings our method obtains clear gains compared with
the state-of-the-art methods.

\section{Preliminaries}

The expected reward $\mtexpect\left(r\mid s,a\right)$ is a core concept
in the contextual bandit to evaluate the policy for the dynamic system.
In case of large state or action spaces, the parameterized approximation
is widely accepted: $\mtexpect\left(r\mid s,a;\mtbfw\right)=\mtbfx\left(s,a\right)^{T}\mtbfw$,
which is assumed to be in a low dimensional space, where $\mtbfw\in\mtbbR^{u}$
is the unknown coefficients and $\mtbfx\left(s,a\right)$ is the contextual
feature for the state-action $\left\{ s,a\right\} $ pair. 

The aim of the actor-critic algorithm is to learn an optimal policy
to maximize the reward for all the state-action pairs. The objective
is $\pi_{\theta^{*}}\!=\!\arg\max_{\theta}\widehat{J}\!\left(\theta\right)$,
where $\widehat{J}\left(\theta\right)=\sum_{s\in\mtcalS}d\left(s\right)\sum_{a\in\mtcalA}\pi_{\theta}\left(a\mid s\right)\mtexpect\left(r\mid s,a;\mtbfw\right)$
is the average reward over all the possible states \& actions; $d\left(s\right)$
is a reference distribution over states. To make the actor updating
a well-posed objective, various constraints on $\theta$ are considered\ \cite{huitian_2014_NIPS_ActCriticBandit4JITAI}.
Specifically, the stochasticity constraint is introduced to reduce
the habituation and facilitate learning\ \cite{huitian_2016_PhdThesis_actCriticAlgorithm}.
The stochasticity constraint specifies the probability of selecting
both actions is at least $p_{0}$ for more than $100\left(1-\alpha\right)\%$
contexts: $P\left[p_{0}\leq\pi_{\theta}\left(a=1\mid s_{t}\right)\leq1-p_{0}\right]\geq1-\alpha$.
Via the Markov inequality, a relaxed and smoother stochasticity constraint
is as follows $\theta^{\intercal}\mtexpect\left[g\left(s\right)^{\intercal}g\left(s\right)\right]\theta\leq\alpha\left\{ \log\left[p_{0}/\left(1-p_{0}\right)\right]\right\} ^{2}$\ \cite{huitian_2016_PhdThesis_actCriticAlgorithm},
leading to the objective $\widehat{J}\left(\theta\right)$ as 
\begin{align}
\widehat{J}\left(\theta\right) & =\sum_{s\in\mtcalS}d\left(s\right)\sum_{a\in\mtcalA}\pi_{\theta}\left(a\mid s\right)\mtexpect\left(r\mid s,a;\mtbfw\right)-\lambda\theta^{\intercal}\mtexpect\left[g\left(s\right)g\left(s\right)^{\intercal}\right]\theta,\label{eq:obj_actorUpdating_LeiPhdThesis}
\end{align}
where $g\left(s_{i}\right)=g\left(s_{i},1\right)-g\left(s_{i},0\right)$;
$g\left(s,a\right)$ is the feature for the policy\ \cite{huitian_2016_PhdThesis_actCriticAlgorithm}. 

According to\ \eqref{eq:obj_actorUpdating_LeiPhdThesis}, we need
the estimation of the expected reward to form the objective. This
process is called the critic updating\ \cite{Grondman_2012_IEEEts_surveyOfActorCriticRL}.
Current methods generally use the ridge regression to learn it. The
objective is defined as follows 
\begin{equation}
\min_{\mtbfw}\sum_{i=1}^{T}\left\Vert \mtbfx\left(s_{i},a_{i}\right)^{\intercal}\mtbfw-r_{i}\right\Vert _{2}^{2}+\zeta\left\Vert \mtbfw\right\Vert _{2}^{2}.\label{eq:obj_criticUpdating_current}
\end{equation}
It has a closed-form solution: $\mhbfw=\left(\mtbfX\mtbfX^{\intercal}+\zeta\mtbfI_{u}\right)^{-1}\mtbfX\mtbfr$,
where $\mtbfX\in\mtbbR^{u\times T}$ is a designed matrix with the
$i$-th column as $\mtbfx_{i}=\mtbfx\left(s_{i},a_{i}\right)$; $\mtbfr=\left[r_{1},\cdots,r_{T}\right]^{\intercal}\in\mtbbR^{T}$
consists of all the immediate rewards. However similar to the existing
least square based algorithms, the objective is sensitive to the existence
of outliers\ \cite{fpNie_2012_ICDM_robustMatrixCompletion,fpNie_2010_NIPS_JointL21_featureSelection}.

\section{Robust Contextual Bandit with Capped-$\ell_{2}$ norm}

To boost the robustness of the actor-critic learning, the capped-$\ell_{2}$
norm is used to measure the approximation error: 
\begin{equation}
\min_{\mtbfw}\ O\left(\mtbfw\right)=\sum_{i=1}^{M}\min\left\{ \left\Vert r_{i}-\mtbfx_{i}^{T}\mtbfw\right\Vert _{2}^{2},\epsilon\right\} +\zeta\left\Vert \mtbfw\right\Vert _{2}^{2}.\label{eq:obj_expectedRwd_cappedL2}
\end{equation}
By properly setting the value of $\epsilon$, we can get rid of the
outliers that distribute far away from the majority of samples while
keep the effective samples. Otherwise when $\epsilon$ is too large,
there are outliers left in the data; while $\epsilon$ is too small,
lots of effective samples will be removed, leading to unstable estimations. 

It is important to properly set the value of $\epsilon$. We propose
an effective method to set $\epsilon$. It is derived from one of
the most widely accepted outlier definitions in the statistics community.
When we use the boxplot to give a descriptive illustration of the
distribution of a dataset, the samples that are $1.5\times IQR$ more
above the third quartile are treated as outliers. Thus, we set $\epsilon$
as:
\begin{equation}
\epsilon=\tau\left(q_{3}+1.5\times IQR\right)\label{eq:outlierDetection_boxplot}
\end{equation}
 where $IQR=q_{3}-q_{1}$ is the interquartile range; $\tau$ is a
tuning parameter to give us a flexible setting of $\epsilon$, which
is set to $1$ by default.

\subsection{\label{sub:Alg_4_ourCriticUpdating}Algorithm for the Critic Updating}
\begin{proposition}
\label{prop:objective_simplication}The critic objective\ \eqref{eq:obj_expectedRwd_cappedL2}
is equivalent to the following objective 
\begin{equation}
\min_{\mtbfw}\sum_{i}u_{i}\left\Vert r_{i}-\mtbfx_{i}^{T}\mtbfw\right\Vert _{2}^{2}+\zeta\left\Vert \mtbfw\right\Vert _{2}^{2},\label{eq:obj_cappedL2_general_simplified-1}
\end{equation}
where $u_{i}=1_{\left\{ \left\Vert r_{i}-\mtbfx_{i}^{T}\mtbfw\right\Vert _{2}^{2}<\epsilon\right\} }$
is dependent on the unknown variable $\mtbfw$. 
\end{proposition}

According to Proposition\ \ref{prop:objective_simplication}, we
have a simplified objective for the critic updating. However, it is
still complex to minimize\ \eqref{eq:obj_cappedL2_general_simplified-1}
since the weight term depends on the unknown variable $\mtbfw$. In
this section, an iteratively re-weighted algorithm is proposed for
the optimization of\ \eqref{eq:obj_cappedL2_general_simplified-1}
(cf. Algorithm\ \ref{alg:2_actCritic_RobustContextualBandit}). It
assumes the weight $\mtbfu$ is fixed when seeking for the optimal
$\mtbfw$ and vice versa. When $\mtbfu$ is fixed, the objective\ \eqref{eq:obj_cappedL2_general_simplified-1}
is convex over $\mtbfw.$ We may get the solver by differentiating\ \eqref{eq:obj_cappedL2_general_simplified-1}
and setting the derivative to zero, leading to the following linear
system
\begin{equation}
{\displaystyle \mtbfw^{\left(t\right)}=\left(\mtbfX\mtbfU^{\left(t-1\right)}\mtbfX^{\intercal}+\zeta\mtbfI\right)^{-1}\mtbfX\mtbfU^{\left(t-1\right)}\mtbfr},\label{eq:updatingRule_4_expectedRwd}
\end{equation}
where $\mtbfU^{\left(t-1\right)}=\mtdiag$$\left(\mtbfu^{\left(t-1\right)}\right)$
is the weight at the $\left(t-1\right)$-th iteration. Then we update
the weight term as $u_{i}^{\left(t\right)}=1_{\left\{ \left\Vert r_{i}-\mtbfx_{i}^{T}\mtbfw^{\left(t\right)}\right\Vert _{2}^{2}<\epsilon\right\} }$
for $i=1,\cdots,T$.

\subsection{Algorithm for the Actor Updating}

Since the distribution of $d\left(s\right)$ in the objective\ \eqref{eq:obj_actorUpdating_LeiPhdThesis}
is generally unavailable, we consider the $T$-trial based objective
as follows {\small{}
\begin{equation}
\widehat{J}\left(\theta\right)=\frac{1}{T}\sum_{i=1}^{T}\sum_{a\in\mtcalA}u_{i}\pi_{\theta}\left(a\mid s_{i}\right)\mtexpect\left(r\mid s_{i},a;\mtbfw\right)-\lambda\theta^{\intercal}\left[\frac{1}{T}\sum_{i=1}^{T}u_{i}g\left(s_{i}\right)g\left(s_{i}\right)^{\intercal}\right]\theta,\label{eq:obj_actorUpdate_our}
\end{equation}
}where $\left\{ u_{i}\right\} _{i=1}^{T}$ is the weighted term learned
from the critic updating, cf.\ Section\ \eqref{sub:Alg_4_ourCriticUpdating}.
With the weight $\left\{ u_{i}\right\} _{i=1}^{T}$, the outlier tuples
that have large approximation errors are removed for the actor updating.
As a result, the robustness is boosted. The actor updating aims to
maximizes the objective\ \eqref{eq:obj_actorUpdate_our} over $\theta$.
We use the Sequential Quadratic Programming (SQP) algorithm for the
optimization. Specially, the implementation of SQP with finite-difference
approximation to the gradient in \textsc{\small{}fmincon} is utilized
in our algorithm (cf. Algorithm\ \ref{alg:2_actCritic_RobustContextualBandit}).
\begin{algorithm}[t]
\caption{Robust actor-critic contextual bandit (RS-ACCB).\label{alg:2_actCritic_RobustContextualBandit}}
 \textbf{Input}: $\zeta,\lambda,\epsilon,\mtbfu=\left[1,\cdots,1\right]\in\mtbbR^{T}$

\begin{algorithmic}[1] 

\STATE Initialize the state $s_{0}\in\mtbbR^{p}$ and policy parameters. 

\REPEAT 

\STATE \emph{/{*}Critic updating to estimate the expected reward
$\mtexpect\left(r\mid s,a;\mtbfw\right)${*}/}

\REPEAT

\STATE Update the parameter for the expected reward $\mbox{\ensuremath{\mhbfw}\ for via\ }$\eqref{eq:updatingRule_4_expectedRwd}.

\STATE Update the weight term $\mtbfu$ according to the estimated
$\mhbfw$.

\UNTIL{convergence }

\STATE \emph{Actor updating} to estimate the policy parameter $\widehat{\theta}=\arg\max_{\theta}\widehat{J}\left(\theta\right)$,
where $\widehat{J}\left(\theta\right)$ is defined in\ \eqref{eq:obj_actorUpdate_our}.

\UNTIL{convergence }

\end{algorithmic} 

\textbf{Output}: the stochastic policy, i.e. $\pi_{\widehat{\theta}}\left(a\mid s\right)$.
\end{algorithm}

\section{Experiments}

\subsection{Datasets}

To evaluate the performance, we utilize a dataset from the mHealth
study (called HeartSteps) to approximate the generative model. The
HeartSteps is a 42-day mHealth study, resulting in 210 decision points
per user. It aims to increase the users' daily activities (i.e. steps)
by sending them positive interventions, for example, suggesting them
to go for a hike on the sunny weekend etc. 

For each user, a trajectory of $T\!=\!210$ tuples of observations
$\mtcalD\!=\!\left\{ \left(s_{i},a_{i,}r_{i}\right)\right\} _{i=1}^{T}$
are generated via the micro-randomized trials\ \cite{SusanMurphy_2016_CORR_BatchOffPolicyAvgRwd,PengLiao_2015_Proposal_offPolicyRL}.
The initial state is drawn from the Gaussian distribution $S{}_{0}\sim\mtcalN_{p}\left\lbrace 0,\Sigma\right\rbrace $,
with the pre-defined covariance matrix $\Sigma\in\mtbbR^{p\times p}$.
The random policy provides a method to select actions. $\forall t\geq0,$
$a_{t}=1$ is chosen with a probability of $0.5$, i.e. $\mu\left(1\mid s_{t}\right)=0.5$
for all states $s_{t}$. When $t\geq1$, the state and immediate
reward are generated as 
\begin{align}
S_{t,1}=\  & \beta_{1}S_{t-1,1}+\xi_{t,1},\nonumber \\
S_{t,2}=\  & \beta_{2}S_{t-1,2}+\beta_{3}A_{t-1}+\xi_{t,2},\label{eq:Dat=0000231_stateTrans_cmp3}\\
S_{t,3}=\  & \beta_{4}S_{t-1,3}+\beta_{5}S_{t-1,3}A_{t-1}+\beta_{6}A_{t-1}+\xi_{t,3},\nonumber \\
S_{t,j}=\  & \beta_{7}S_{t-1,j}+\xi_{t,j},\qquad\mtfor\ j=4,\ldots,p\nonumber \\
R_{t}=\  & \beta_{14}\times\left[\beta_{8}+A_{t}\times\left(\beta_{9}+\beta_{10}S_{t,1}+\beta_{11}S_{t,2}\right)+\beta_{12}S_{t,1}-\beta_{13}S_{t,3}+\varrho_{t}\right],\label{eq:Dat=0000231_ImmediateRwd_cmp3}
\end{align}
where $\bm{\beta}=\left\{ \beta_{i}\mid i=1,\cdots,14\right\} $ is
the main coefficient for the dynamic system. It is set as $\bm{\beta}=${[}0.4,0.3,0.4,0.7,0.05,0.6,
0.25, 3,0.25,0.25,0.4,0.1,0.5,500{]}. $\left\{ \xi_{t,i}\right\} _{i=1}^{p}\sim\mtcalN\left(0,\sigma_{s}^{2}\right)$
is the Gaussian noise in the state\ \eqref{eq:Dat=0000231_stateTrans_cmp3}
and $\varrho_{t}\sim\mtcalN\left(0,\sigma_{r}^{2}\right)$ is the
Gaussian noise in the reward model\ \eqref{eq:Dat=0000231_ImmediateRwd_cmp3}. 

To simulate the outliers in the trajectory, there are two processing
steps: (a) a fixed ratio (i.e. $\psi=4\%$) of tuples is randomly
selected in each user's trajectory; (b) we add a large noise ($\nu$
times the average value in the trajectory) to the states and rewards
in the selected tuples. Additionally, the actions in the selected
tuples are randomly set to simulate the random failure of sending
interventions due to the weak mobile network.

\subsection{Experiments Settings }

In the experiment, there are three contextual bandit methods for comparison:
(1) Lin-UCB (linear upper confidence bound) is a famous contextual
bandit method that achieves great successes in the Internet advertising\ \cite{YihongLi_2010_WWW_contextualBandit4newsArticleRecommend,lihong_2011_ICML_doublyRobust,Ambuj_2017_Springer_FromAds_Interventions};
(2) S-ACCB is the stochasticity constrained actor-critic contextual
bandit for the mHealth\ \cite{huitian_2016_PhdThesis_actCriticAlgorithm};
(3) RS-ACCB is the proposed Robust ACCB with the stochasticity constraint.

We use the the expected long-run average reward (ElrAR)\ \cite{SusanMurphy_2016_CORR_BatchOffPolicyAvgRwd}
to evaluate the estimated policies $\pi_{\widehat{\theta}_{n}}$ for
$n\in\left\{ 1,\cdots,N\right\} $. There are two processing steps
to obtain the ElrAR: (a) get the average reward $\eta^{\pi_{\widehat{\theta}_{n}}}$
for the $n$-th user by averaging the rewards over the last $4,000$
elements in a trajectory of $5,000$ tuples under the policy $\pi_{\widehat{\theta}_{n}}$;
(b) the ElrAR $\mathbb{E}\left[\eta^{\pi_{\hat{\theta}}}\right]$
is achieved by averaging the $50$ $\eta^{\pi_{\widehat{\theta}}}$'s.

There are $N=50$ users' MDPs used in the experiment. Each user has
a trajectory of $T=210$ tuples. There are $p=3$ variables in the
state. The noises in the MDP are set as $\sigma_{r}=3$ and $\sigma_{s}=1$
respectively. The parameterized policy is assumed to be the Boltzmann
distribution $\pi_{\theta}\left(a\mid s\right)\!=\!\frac{\exp\left[-\theta^{\intercal}g\left(s,a\right)\right]}{\sum_{a'}\exp\left[-\theta^{\intercal}g\left(s,a'\right)\right]}$\ \cite{SusanMurphy_2016_CORR_BatchOffPolicyAvgRwd},
where $\theta\in\mtbbR^{m}$ is the unknown coefficients, $g\left(s,a\right)=\left[as^{\intercal},a\right]^{\intercal}$
is the policy feature and $m=p+1$. The feature vector for the estimation
of expected rewards is set as $\mtbfx\left(s,a\right)=\left[1,s^{\intercal},a,s^{\intercal}a\right]^{\intercal}\in\mtbbR^{u}$,
where $u=2p+2$. The tuning parameters for the actor-critic learning
are set as $\zeta=\lambda=0.001$. The outlier ratio and strength
are set $\psi=4\%$ and $\nu=5$ respectively. In our algorithm, $\tau$
is set as 1. 
\begin{figure}[t]
\begin{centering}
\includegraphics[width=0.95\linewidth]{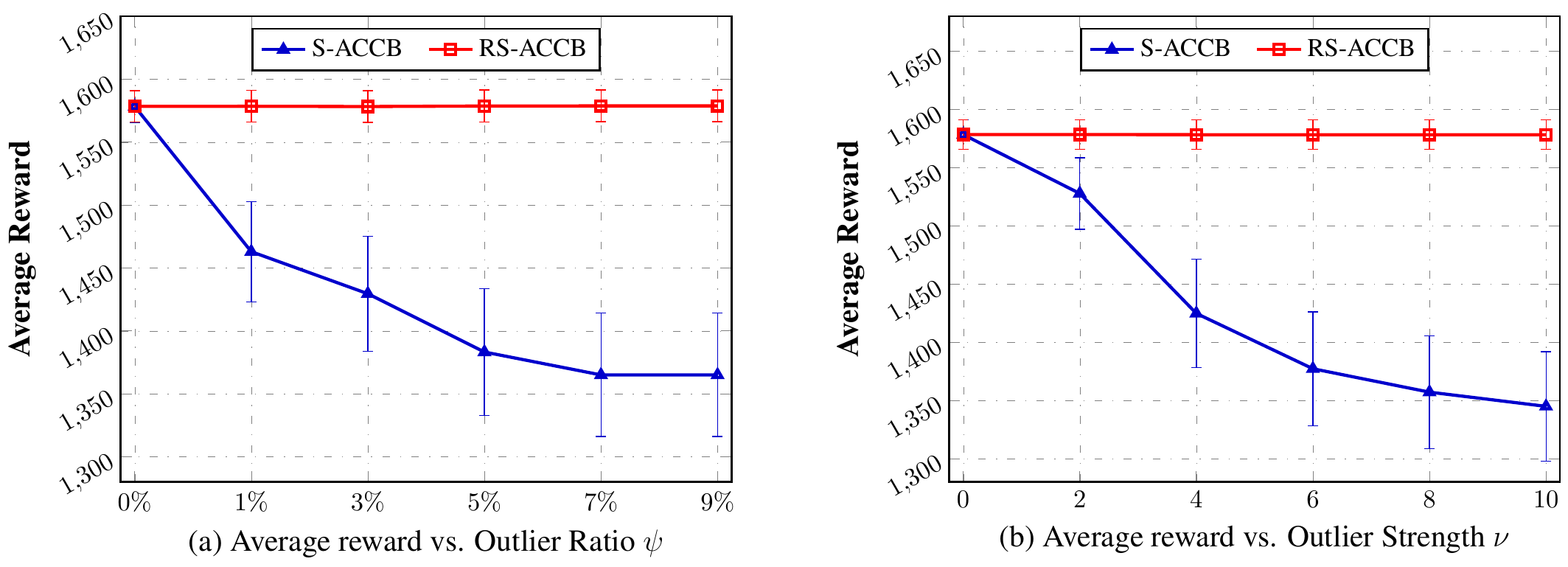}\caption{Average reward of three contextual bandit methods. The left sub-table
shows the results when the trajectory is short, i.e. $T=42$; the
right one shows the results when $T=100$. RS-ACCB is our method.
A larger value is better. \label{fig:P_vs_outlierRatioStrength}}

\par\end{centering}

\vspace{-0.5cm}
\end{figure}

\subsection{Results and Discussion\label{sub:EvaluationMetric-and-Results} }

In this section, the experiments are carried out to verify the performance
of three contextual bandit methods from the following two aspects:

(\textbf{S1}) We change the ratio $\psi$ of tuples that contain outliers
from $0\%$ to $9\%$. The experiment results are displayed in the
left sub-table in Table\ \ref{tab:AverageRwd_PooledRL_SeparRL_CD-RL}
and Fig.\ \ref{fig:P_vs_outlierRatioStrength}(a).  As we can see,
when $\psi=0$, there is zero percentage of outliers in the dataset.
Under such condition, our method achieves almost identical results
compared with the S-ACCB\ \cite{huitian_2016_PhdThesis_actCriticAlgorithm}.
This results verify that though our method aims to the robust learning,
it is well adapted to the dataset without outliers. As $\psi$ rises,
the performance of both Lin-UCB and S-ACCB drops obviously. While
their standard deviations increases dramatically. Compared with those
two methods, both the performance and the standard deviation of our
method keep stable. As a result, our method averagely improves the
performance by 146.8 steps, i.e. 10.26\%, compared with the best of
state-of-the-art methods.

(\textbf{S2}) The strength of outliers $\nu$ ranges from $0$ to
$10$ times of the average value in the trajectory. The right sub-table
in Table\ \ref{tab:AverageRwd_PooledRL_SeparRL_CD-RL} and Fig.\ \ref{fig:P_vs_outlierRatioStrength}(b)
summarize the experiment results. As we shall see, when $\nu$ rises,
the strength of outliers increases gradually. We have the following
observations from the experiment results: (1) when there is no outlier
in the trajectory, our method achieves similar results compared with
S-ACCB; (2) as $\nu$ rises, the performances of S-ACCB and Lin-UCB
decrease obviously and their standard deviations increase dramatically;
(3) as $\nu$ rises, both the performance of our method and the standard
deviation keep stable. Compared with the state-of-the-art methods,
our method get clear gains in a variety of parameter settings. Averagely,
it improves the performance by 139.3 steps and 143.3 steps compared
with Lin-UCB and S-ACCB respectively. 
\begin{table}[t]
\begin{centering}
\caption{Average reward vs. ourlier ratio $\psi$ (setting \textbf{S1}) and
outlier strength $\nu$ (setting \textbf{S2}) on the two sub-tables.
The three methods are (a) Lin-UCB\ \cite{YihongLi_2010_WWW_contextualBandit4newsArticleRecommend},
(b) S-ACCB\ \cite{huitian_2016_PhdThesis_actCriticAlgorithm} and
(c) RS-ACCB (is our method). A larger value is better.\label{tab:AverageRwd_PooledRL_SeparRL_CD-RL}}

\par\end{centering}

\begin{adjustbox}{width=1\linewidth, center}

\begin{centering}
\begin{tabular}{|c|ccc|c|ccc|}
\hline 
\multirow{2}{*}{$\psi$} &
\multicolumn{3}{c|}{Average reward vs. outlier ratio $\psi$} &
\multirow{2}{*}{$\nu$} &
\multicolumn{3}{c|}{Average reward vs. outlier strength $\nu$}\tabularnewline
\cline{2-4} \cline{6-8} 
 & Lin-UCB &
S-ACCB &
RS-ACCB &
 & Lin-UCB &
S-ACCB &
RS-ACCB\tabularnewline
\hline 
$0\%$ &
\textit{\textcolor{blue}{1578.7$\pm$13.75}} &
1578.3$\pm$12.70 &
1578.3$\pm$12.55 &
$0$ &
\textit{\textcolor{blue}{1578.7$\pm$13.75}} &
1578.3$\pm$12.70 &
1578.3$\pm$12.55\tabularnewline
$1\%$ &
1462.5$\pm$40.24 &
1462.9$\pm$39.88 &
\textcolor{blue}{\emph{1578.4$\pm$12.61}} &
$2$ &
1535.6$\pm$21.94 &
1527.7$\pm$30.71 &
\textcolor{blue}{\emph{1578.3$\pm$12.68}}\tabularnewline
$3\%$ &
1428.1$\pm$49.69 &
1429.5$\pm$45.79 &
\textcolor{blue}{\emph{1578.2$\pm$12.57}} &
$4$ &
1431.7$\pm$44.13 &
1424.7$\pm$46.53 &
\textcolor{blue}{\emph{1578.2$\pm$12.65}}\tabularnewline
$5\%$ &
1391.0$\pm$49.42 &
1383.2$\pm$50.40 &
\textcolor{blue}{\emph{1578.6$\pm$12.66}} &
$6$ &
1380.8$\pm$49.03 &
1377.2$\pm$48.83 &
\textcolor{blue}{\emph{1578.2$\pm$12.62}}\tabularnewline
$7\%$ &
1370.6$\pm$50.20 &
1365.0$\pm$49.02 &
\textcolor{blue}{\emph{1578.7$\pm$12.62}} &
$8$ &
1359.8$\pm$49.76 &
1357.1$\pm$48.51 &
\textcolor{blue}{\emph{1578.2$\pm$12.63}}\tabularnewline
$9\%$ &
\multicolumn{1}{c}{1358.9$\pm$48.43} &
1365.0$\pm$49.02 &
\textcolor{blue}{\emph{1578.7$\pm$12.62}} &
$10$ &
1346.8$\pm$48.83 &
1344.9$\pm$46.94 &
\textcolor{blue}{\emph{1578.2$\pm$12.64}}\tabularnewline
\hline 
Avg &
1431.6 &
1430.7 &
\textcolor{blue}{\emph{1578.5}} &
Avg &
1438.9 &
1435.0 &
\textcolor{blue}{\emph{1578.2}}\tabularnewline
\hline 
\end{tabular} 
\par\end{centering}

\end{adjustbox}

\vspace{-0.5cm}
\end{table}

\section{Conclusions and Future Directions}

To alleviate the influence of outliers in the mHealth study, a robust
actor-critic contextual bandit method is proposed to form robust interventions.
We use the capped-$\ell_{2}$ norm to boost the robustness for the
critic updating, which results in a set of weights. With them, we
propose a weighted objective for the actor updating. It gives the
tuples that have large approximate errors zero weights, enhancing
the robustness against those tuples. Additionally, a solid method
is provided to properly set the thresholding parameter in the capped-$\ell_{2}$
norm, i.e., $\epsilon.$ With it, we can achieve the conflicting goal
of enhancing the robustness of the actor-critic algorithm as well
as obtaining almost identical results compared with the state-of-the-art
method on the datasets without outliers. Extensive experiment results
show that in a variety of parameter settings the proposed method obtains
significant improvements compared with the state-of-the-art contextual
bandit methods. In the future, we may explore the robust learning
on the reinforcement learning method. It could be on both the discount
reward setting and the average reward setting\ \cite{Grondman_2012_IEEEts_surveyOfActorCriticRL,SusanMurphy_2016_CORR_BatchOffPolicyAvgRwd}.
Those two directions are much more challenging since it is not a general
regression task to estimate the value function. Besides, mining the
cohesion information among users helps a lot to enrich the data (or
restrict the parameter space)\ \cite{fyzhu_2014_IJPRS_SSNMF,haichangLi_2016_IJRS_LablePropagationHyperClassification,guangliangCheng_2014_ICIP,guangliangCheng_2015_ICIP,guangliangCheng_2016_neurocomputing,guangliangCheng_2016_TGRSL}.

\section*{Appendix: the proof of Proposition\ \ref{prop:objective_simplication}}
\begin{proof}
The objective of\ \eqref{eq:obj_expectedRwd_cappedL2} is non-convex
and non-differentiable\ \cite{QianSun_2013_SIGKDD_CappedNorm4RPCA,hongchangGao_2015_CIKM_cappedNorm4NMF}.
We could obtain its sub-gradient: $\partial O\left(\mtbfw\right)=\sum_{i}\partial\min\left\{ \left\Vert r_{i}-\mtbfx_{i}^{T}\mtbfw\right\Vert _{2}^{2},\epsilon\right\} +2\zeta\mtbfw,$
where {\small{}
\begin{equation}
\partial\min\left\{ \left\Vert r_{i}-\mtbfx_{i}^{T}\mtbfw\right\Vert _{2}^{2},\epsilon\right\} =\begin{cases}
0, & \mtif\ \left\Vert r_{i}-\mtbfx_{i}^{T}\mtbfw\right\Vert _{2}^{2}>\epsilon\\
\left[-1,0\right]\partial\left(\left\Vert r_{i}-\mtbfx_{i}^{T}\mtbfw\right\Vert _{2}^{2}\right) & \mtif\ \ r_{i}-\mtbfx_{i}^{T}\mtbfw=-\sqrt{\epsilon}\\
\left[0,1\right]\partial\left(\left\Vert r_{i}-\mtbfx_{i}^{T}\mtbfw\right\Vert _{2}^{2}\right) & \mtif\ \ r_{i}-\mtbfx_{i}^{T}\mtbfw=\sqrt{\epsilon}\\
\partial\left(\left\Vert r_{i}-\mtbfx_{i}^{T}\mtbfw\right\Vert _{2}^{2}\right) & \mtif\ \left\Vert r_{i}-\mtbfx_{i}^{T}\mtbfw\right\Vert _{2}^{2}<\epsilon
\end{cases}.\label{eq:sub_gradient}
\end{equation}
}Letting $u_{i}=1_{\left\{ \left\Vert r_{i}-\mtbfx_{i}^{T}\mtbfw\right\Vert _{2}^{2}<\epsilon\right\} }$
for $i\in\left\{ 1,\cdots,T\right\} $ gives a simplified partial
derivative of\ \eqref{eq:obj_expectedRwd_cappedL2} that satisfies
the sub-gradient\ \eqref{eq:sub_gradient}. It is defined as 
\[
\partial O\left(\mtbfw\right)=\sum_{i}u_{i}\partial\left(\left\Vert r_{i}-\mtbfx_{i}^{T}\mtbfw\right\Vert _{2}^{2}\right)+2\zeta\mtbfw,
\]
which is equivalent to the partial derivative of the following objective
\begin{equation}
\max_{\mtbfw}\sum_{i}u_{i}\left\Vert r_{i}-\mtbfx_{i}^{T}\mtbfw\right\Vert _{2}^{2}+\zeta\left\Vert \mtbfw\right\Vert _{2}^{2}.\label{eq:obj_cappedL2_general_simplified}
\end{equation}
From the perspective of optimization, the objective\ \eqref{eq:obj_cappedL2_general_simplified}
is equivalent to\ \eqref{eq:obj_expectedRwd_cappedL2}. {\small{}\bibliographystyle{2_home_fyzhu_DATA_Dropbox_self_Folder_myWorksOn___CCAI_RobustRL_4_mHealth_onlineLearning_ieee}
\bibliography{1_home_fyzhu_link2dropbox_self_Folder_myWorksOnDropboxs_bibFiles_referenceBib2,3_home_fyzhu_link2dropbox_self_Folder_myWorksOnDropboxs_bibFiles_referenceBib}
}{\small \par}\end{proof}

\end{document}